\documentclass[wcp]{jmlr}

\usepackage{longtable}

\usepackage{booktabs} 
\usepackage{xspace}
\usepackage{listings}
\usepackage{algorithm}
\usepackage{algorithmic}
\usepackage{amsmath}
\usepackage{bbm}
\usepackage{adjustbox}
\usepackage{bm}
\usepackage{hyperref}
\usepackage{amsmath}
\usepackage{amssymb}
\usepackage{makecell}
\usepackage{wrapfig}

\makeatletter
\let\Ginclude@graphics\@org@Ginclude@graphics 
\makeatother

\newcommand{\revision}[1]{#1}
\newcommand{\method}{MARC\xspace}
\newcommand{\rulesep}{\unskip\ \vrule\ }

\jmlrvolume{XX}
\jmlryear{2022}
\jmlrworkshop{ACML 2022}

\title[Margin Calibration for Long-Tailed Visual Recognition]{Margin Calibration for Long-Tailed Visual Recognition}

  \author{\Name{Yidong Wang}$^{1}$\thanks{Equal contribution.}, \Name{Bowen Zhang}$^{1*}$, \Name{Wenxin Hou}$^{2}$, \Name{Zhen Wu}$^{3}$\thanks{Corresponding to wuz@nju.edu.cn, jindong.wang@microsoft.com.}, \Name{Jindong Wang}$^{4\dagger}$,  \Name{Takahiro Shinozaki}$^{1}$ \\
  \addr $^{1}$Tokyo Institute of Technology, $^{2}$Microsoft STCA, $^{3}$Nanjing University, $^{4}$Microsoft Research Asia
 }

\jmlrvolume{189}
\editors{Emtiyaz Khan and Mehmet G\"{o}nen}

\begin{document}

\maketitle

\begin{abstract}
\revision{Long-tailed visual recognition tasks} pose great challenges for neural networks on how to handle the imbalanced predictions between head \revision{(common)} and tail \revision{(rare)} classes, i.e., models tend to classify tail classes as head classes. While existing research focused on data resampling and loss function engineering, in this paper, we take a different perspective: the \emph{classification margins}. We study the relationship between the margins and \revision{logits} and empirically observe that the \revision{uncalibrated} margins and logits are \emph{positively correlated}. We propose a simple yet effective \revision{ \emph{MARgin Calibration} approach (\textbf{MARC})} to \revision{
calibrate} the margins to obtain better logits. We validate \revision{MARC} through extensive experiments on common long-tailed benchmarks including CIFAR-LT, ImageNet-LT, Places-LT, and iNaturalist-LT. Experimental results demonstrate that our \revision{MARC} achieves favorable results on these benchmarks. In addition, \revision{MARC} is extremely easy to implement with just three lines of code. We hope this simple approach will motivate people to rethink the \revision{uncalibrated} margins and logits in long-tailed visual recognition.
\end{abstract}


\section{Introduction}
Despite the great success of neural networks in the visual recognition field~\citep{simonyan2014very,he2016deep}, it is still challenging for neural networks to deal with the ubiquitous long-tailed datasets in the real world~\citep{yang2022survey,buda2018systematic,kang2019decoupling,zhou2020bbn}. To be clear, in the long-tailed datasets, the high-frequency classes \revision{(head/common classes)} occupy most of the instances, whereas the low-frequency classes \revision{(tail/rare classes)} involve a small amount of instances~\citep{liu2019large,van2017devil}. Due to the imbalance of training data, the model performs well in head classes with much worse performance in tail classes~\citep{buda2018systematic,zhang2021distribution}.

Towards addressing the long-tailed recognition problem, there are several strategies such as data re-sampling and loss function engineering.
Data re-sampling aims to `simulate' a balanced training dataset by over-sampling the tail class or under-sampling the head classes~\citep{ando2017deep,buda2018systematic,pouyanfar2018dynamic,shen2016relay}, while loss re-weighting is introduced to adjust the weights of losses for different classes or different instances~\citep{byrd2019effect,khan2017cost,wang2017learning}. For more balanced gradients between classes, some class-balanced loss functions adjust the logits instead of weighting the losses~\citep{menon2020long,cao2019learning,ren2020balanced}.\looseness=-1

However, as pointed out by existing research~\citep{ganganwar2012overview,zhou2005training,cao2019learning}, data re-sampling strategies and loss re-weighting schemes will possibly cause underfitting on the head class and overfitting on the tail class. On the other hand, class-balanced loss functions or data re-sampling will lead to worse data representations compared with the standard training using the cross-entropy loss and the instance-balanced sampling (i.e., each instance has the same probability of being sampled)~\citep{kang2019decoupling,ren2020balanced}. In addition, \revision{recent research} reveals that the \revision{uncalibrated} decision boundary given by the classifier head seems to be the performance bottleneck of the long-tailed visual recognition~\citep{kang2019decoupling,zhang2021distribution}. To benefit from both good data representations and the unbiased decision boundary, Decoupling, a heuristic two-stage strategy is proposed to adjust the initially-learned classifier head~\citep{kang2019decoupling} after the standard training. Furthermore, distribution alignment~\citep{zhang2021distribution} is developed as an adaptive calibration function to adjust the initially trained \revision{logits} for each data point. However, as pointed out by existing research~\citep{platt1999large,elsayed2018large}, the margins and logits have a critical effect to the classification performance. \textit{\revision{The} relation between the \revision{uncalibrated} margin and the logits is neglected in existing research}, where the margin is the distance from the data point to the decision boundary.

\begin{figure}[t!]
     \centering
     \subfigure[Logits]{
        \includegraphics[width=0.27\textwidth]{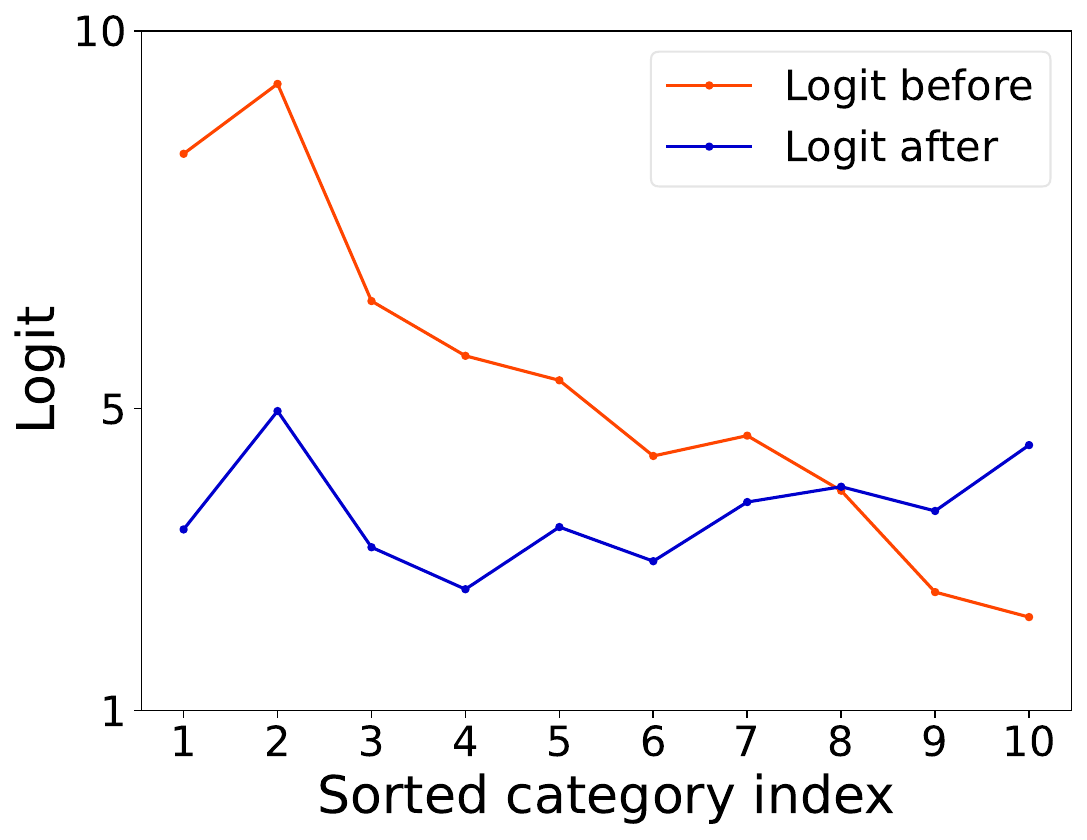}
        \label{fig:logit cifar10}
    }
    \subfigure[Margins]{
    \includegraphics[width=.27\textwidth]{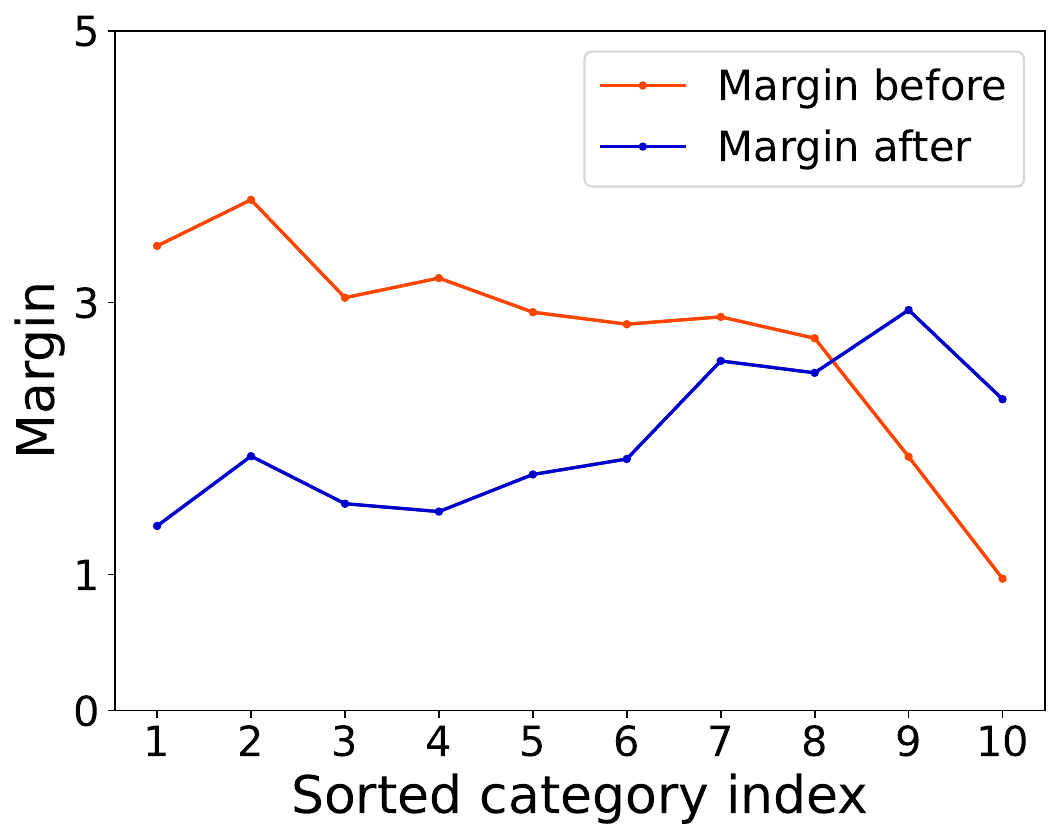}
    \label{fig:margin cifar10}
    }
    \subfigure[Class-wise accuracy]{
         \includegraphics[width=0.37\linewidth]{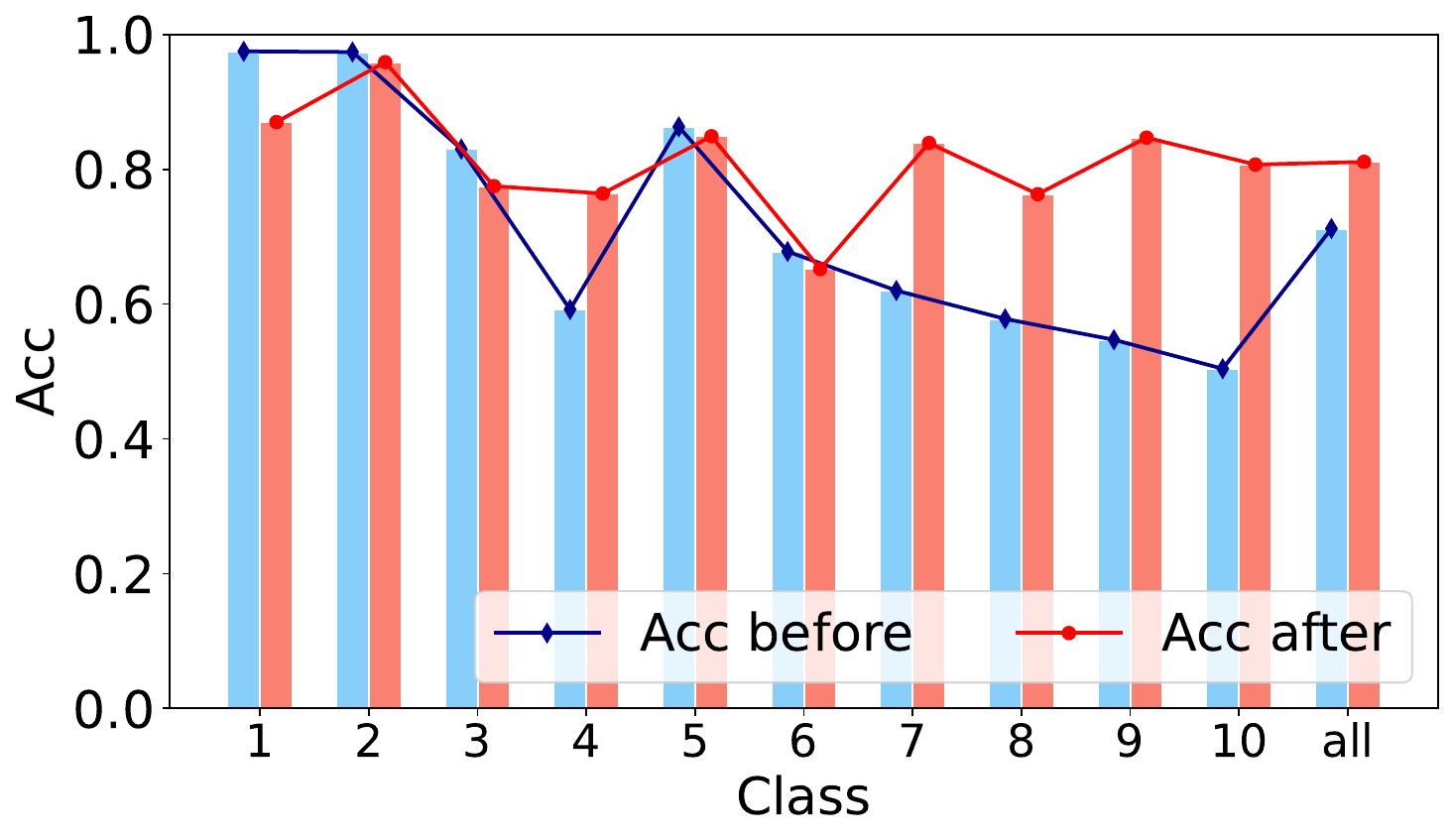}
         \label{fig:cifar10 acc}
         }

    \caption{\textbf{Logits, margins, and class-wise accuracy of CIFAR-10-LT with imbalance factor 200}. Here, \revision{the logit and margin represent the average logit and margin of each class}, \textit{before} and \textit{after} refers to standard training and our method, respectively. The class \revision{indices} are sorted by the number of samples (Head to tail).}
        \label{fig:logits and margins cifar10}
\end{figure}

In this paper, we study the relationship between margins and logits, which are critical factors that dominate the long-tailed performance.
As shown in \figurename~\ref{fig:logits and margins cifar10}, we empirically find that the margin and the logit are correlated with the cardinality of each class. To be concrete, before any calibration, \textit{head classes tend to have much larger margins and logits than tail classes}. Therefore, it is necessary to \revision{calibrate} the margin to obtain the balanced logits. More importantly, as shown in \figurename~\ref{fig:cifar10 acc}, the \revision{uncalibrated} margins and logits will have a negative impact on the classification performance. Therefore, it remains challenging to design an efficient method for such calibration that can achieve satisfying performance without introducing much computational burden.

The model can perform better with the confidence calibration~\citep{platt1999large,guo2017calibration,zhang2021distribution}. however, in this work, we focus on calibrating the biased margins and propose a simple yet effective \textbf{MAR}gin \textbf{C}alibration (\textbf{\revision{\method}}) approach for long-tailed recognition.
In detail, after getting the representations and the classifier head from the standard training, we propose a simple class-specific margin calibration function with only $2 K$ learnable parameters to adjust the initially learned margins, where $K$ is the number of classes. As demonstrated in \figurename~\ref{fig:logits and margins cifar10}, the logits are more balanced when using \revision{\method}.
We conduct experiments on several popular long-tail benchmark datasets: CIFAR-10-LT~\citep{krizhevsky2009learning}, CIFAR-100-LT~\citep{krizhevsky2009learning}, Places-LT~\citep{zhou2017places}, iNaturalist
2018~\citep{van2018inaturalist}, and ImageNet-LT~\citep{liu2019large}. The results demonstrate that our proposed \revision{MARC} approach performs remarkably well while remaining very simple to implement. We hope that our exploration will attract attention to the imbalanced margins in long-tailed recognition.

To sum up, our contributions are as follows:

\begin{itemize}
    \item For the first time in long-tailed recognition, we study the \revision{uncalibrated} predictions from a margin-based perspective. We empirically find that \revision{uncalibrated} margins will cause imperfect predictions, \revision{which could lead to future algorithm designs}.
    
    \item Based on our observations, we propose a simple yet effective margin calibration (\revision{MARC}) approach with only $2K$ trainable parameters to adjust the margins \revision{to get the unbiased prediction for long-tailed visual recognition problem}.
    
    \item \revision{Compared with SOTA methods~\citep{zhang2021distribution,hong2021disentangling}, \revision{\method} is competitive on various long-tailed visual benchmarks} like CIFAR-10-LT~\citep{krizhevsky2009learning}, CIFAR-100-LT~\citep{krizhevsky2009learning}, ImageNet-LT~\citep{liu2019large}, Places-LT~\citep{zhou2017places} and iNaturalist2018~\citep{van2018inaturalist}. In addition, it is extremely easy to implement with just three lines of code.
\end{itemize}

\section{Related Work}
Long-tailed visual recognition has attracted much attention for its commonness in the real world~\citep{yang2022survey,he2009learning,buda2018systematic,kang2019decoupling,ren2020balanced,yang2020rethinking,hong2021disentangling}. Existing methods can be divided into four categories.

\textbf{Data re-sampling.} Data re-sampling \revision{techniques} re-sample the imbalanced training dataset to `simulate' a balanced training dataset. These methods include under-sampling, over-sampling, and classed-balanced sampling. Under-sampling decreases the probability of the instance of head classes being sampled~\citep{drummond2003c4}, whereas over-sampling makes instances of tail classes more likely to be sampled~\citep{chawla2002smote,han2005borderline,wang2021rsg}. Class-aware sampling chooses instances of each class with the same probabilities~\citep{shen2016relay}.

\textbf{Loss function engineering.} Loss function engineering is another direction to obtain balanced gradients during the training. The typical methods can be categorized as loss-reweighting and logits adjustment. Loss re-weighting adjusts the weights of losses for different classes or different instances in a more balanced manner, i.e. the instances in tail classes have larger weights than those in head classes~\citep{byrd2019effect,khan2017cost,wang2017learning}. On the other hand, instead of re-weight losses, some class-balanced loss functions adjust the logits to get balanced gradients during training~\citep{menon2020long,cao2019learning,ren2020balanced,yang2009margin}.

\textbf{Decision boundary adjustment.} Nevertheless, data re-sampling or loss function engineering will influence the representations of data~\citep{ren2020balanced}. Lots of empirical observations show that we can acquire good representation when using the standard training and the classifier head is the performance bottleneck~\citep{kang2019decoupling,zhang2021distribution,yu2020devil,kim2020adjusting}. To solve the above problem, decision boundary adjustment methods re-adjust the classifier head after the standard training \revision{in a learnable way~\citep{kang2019decoupling,zhang2021distribution} or using maximum likelihood estimation such as the Platt scaling~\citep{platt1999large}}. However, they ignore the relationship between the \revision{uncalibrated} margins and logits. \revision{Moreover, \revision{\method} targets a multi-class classification problem and Platting scaling targets binary classification (based on sigmoid), and their updating manners are also different since \revision{\method} adopts an end-to-end manner that updates $2K$ learnable parameters}.

\textbf{Other methods.}
There also exist other paradigms to deal with the long-tailed recognition task, including task-specific \revision{architecture} design~\citep{wang2021contrastive,zhou2020bbn,wang2021rsg}, transfer learning~\citep{liu2019large,yin2019feature}, domain adaptation~\citep{jamal2020rethinking}, semi supervised learning and self supervised learning~\citep{yang2020rethinking}. But these methods either rely on the non-trivial \revision{architecture} design or external data. In contrast, our proposed \revision{MARC} is very simple to implement and does not require external data.
Detailed comparison is shown in \tablename~\ref{tab:diff}.

\section{Method}
\subsection{Preliminaries}

In the popular setting of long-tailed recognition~\citep{kang2019decoupling,cui2019class,ren2020balanced}, the training data distribution is imbalanced while the test data distribution is balanced.
More formally, let $\mathcal{D}=\{(\mathbf{x}_i,y_i)\}_{i=1}^n$ be a training set, where $y_i$ denotes the label of data point $\mathbf{x}_i$.
Specifically, $n = \sum_{j=1}^K n_j$ is the total number of training samples, where $n_j$ is the number of training samples in class $j$ and $K$ is the number of classes.
We assume $n_1 > n_2 > \cdots >n_K$ without loss of generality.
Normally, the prediction function is composed of two modules: the feature representation learning function $f: \mathbf{x} \mapsto \mathbf{z}$ parameterized by $\theta_r$ and the classifier $g: \mathbf{z} \mapsto y$ parameterized by $\theta_c$, where $\mathbf{z} \in \mathbb{R}^{p}$ denotes the feature representation and $p$ is the feature dimension.
Typically, $g$ is a linear classifier that gives the classification score of class $j$ as:
\begin{equation}
    \eta_j = g(\mathbf{z}) := \mathbf{W}_j \mathbf{z} + \mathbf{b}_j,
\end{equation}
where $\mathbf{W}_j$ and $\mathbf{b}_j$ are the weight vector and bias for class $j$, respectively.
Finally, using the softmax function, the probability of $\mathbf{x}_i$ being classified as label $y_i$ is expressed as:
\begin{equation}
\label{eq:softmax}
    p(y=y_i|\mathbf{x}_i;\theta_r,\theta_c) = \frac{\exp(\eta_{y_i})}{\sum_{j=1}^K  \exp(\eta_j)},
\end{equation}
and its loss is computed as the cross-entropy loss:
\begin{equation}
    \label{eq:loss_softmax}
    \ell(\mathbf{x}_i, y_i;\theta_r,\theta_c) = -\log \left(\frac{\exp(\eta_{y_i})}{\sum_{j=1}^K  \exp(\eta_j)} \right).
\end{equation}

\subsection{\revision{Uncalibrated} Margins and Logits}

The decision boundary is often \revision{uncalibrated} in long-tailed recognition, which will lead to imperfect predictions, i.e., the model tends to classify tail classes as head classes. To alleviate this issue, data re-sampling and loss function engineering are two directions to simulate a `balanced' training dataset. However, such techniques will do harm to the representation learning of the model \revision{and lead to low overall accuracy}~\citep{kang2019decoupling,ren2020balanced}. \revision{These methods including ours will sometimes do harm to the performance of head classes. However, the "good" performance of head classes is sometimes not always ensuring positive overall results.} To benefit from both the good representation \revision{that will improve the overall accuracy} and the \revision{calibrated} decision boundary, decision boundary \revision{adjustment} methods are developed~\citep{kang2019decoupling,zhang2021distribution}. However, existing decision boundary adjustment methods ignore such calibration in the margins, which is essential to avoid \revision{un\revision{calibrated}} predictions. Thus, we aim to \revision{calibrate} the margins to obtain balanced predictions.

\begin{wrapfigure}{r}{0.3\textwidth}
\centering
\includegraphics[width=0.3\textwidth]{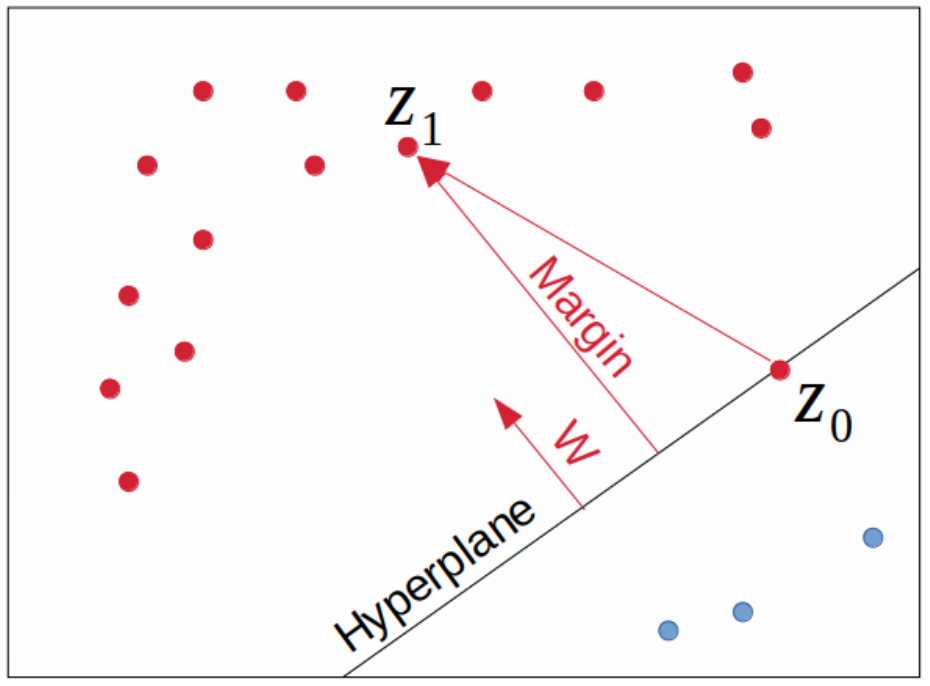}
\caption{\textbf{Illustration of margins.} The red and blue dots denote majority and minority classes, respectively.}
\vspace{-.2in}
\label{fig:margin_intro}
\end{wrapfigure}

In this paper, we find that the \revision{margins~\citep{hastie2009elements}} and logits are biased in long-tailed recognition. The margins are illustrated in \figurename~\ref{fig:margin_intro}.
We define an affine hyperplane $H_j \in \mathbb{R}^{{p}-1} $ of class $j$ as $\mathbf{W}_j \mathbf{z}+\mathbf{b}_j=0$, i.e. any representation point falling on the positive side of $H_j$ can be attributed to class $j$.
Assume that $\mathbf{z}_0$ is a point satisfying $\mathbf{W}_j \mathbf{z}_0 +\mathbf{b}_j=0$, i.e., $\mathbf{z}_0$ is on the hyperplane $H_j$. Suppose $\mathbf{z}_1$ is an arbitrary point in the feature space. We construct the vector $\mathbf{z}_1 - \mathbf{z}_0$ pointing from $\mathbf{z}_0$ to $\mathbf{z}_1$ and project it onto the normal vector $\mathbf{W}_j$. The length of the projection vector $proj_{ \mathbf{W}_j }(\mathbf{z}_1 - \mathbf{z}_0)$ is the margin from $\mathbf{z}_1$ to $H_j$.
More formally, such margin is calculated as:
\begin{equation}
\begin{split}
    d_j & =  \left\Vert  proj_{ \mathbf{W}_j }(\mathbf{z}_1 - \mathbf{z}_0) \right\Vert  \\ 
    & = \left\Vert \frac{ \mathbf{W}_j \cdot (\mathbf{z}_1 - \mathbf{z}_0)}{\mathbf{W}_j \cdot \mathbf{W}_j}  \mathbf{W}_j \right\Vert \\
    & = \frac{\mathbf{W}_j \cdot \mathbf{z}_1 - \mathbf{W}_j \cdot \mathbf{z}_0 }{\Vert  \mathbf{W}_j \Vert } \\
    & =\frac{\mathbf{W}_j \mathbf{z}_1+\mathbf{b}_j}{\Vert \mathbf{W}_j \Vert} \quad (\text{since} \ \mathbf{W}_j  \mathbf{z}_0 + \mathbf{b}_j = 0),
\end{split}
\end{equation}
where $\Vert \cdot \Vert$ denotes L2 norm.
Thus, the logit $\mathbf{W}_j \cdot \mathbf{z}_1 + \mathbf{b}_j$ can also be expressed as $\Vert \mathbf{W}_j \Vert d_j$. Based on this conclusion, we can rewrite \eqref{eq:softmax} as:
\begin{equation}
    \label{eq:dist_softmax}
    p(y=y_i|\mathbf{x}_i;\theta_r,\theta_c) = \frac{\exp(\eta_{y_i})}{\sum_{j=1}^K  \exp(\eta_j)} = \frac{\exp(\Vert \mathbf{W}_{y_i} \Vert d_{y_i})}{\sum_{j=1}^K  \exp(\Vert \mathbf{W}_j \Vert d_j)}.
\end{equation}

Consider a data point is on the decision boundary of class $j$ and class $t$ (on the hyperplane in \figurename~\ref{fig:margin_intro}), i.e., such data point has the same probability of being classified as class $j$ or class $t$. Clearly, the assumed data point on the decision boundary satisfies:
\begin{equation}
    \label{eq:decision_boudary}
     \eta_j = \eta_t = \Vert \mathbf{W}_j \Vert d_j = \Vert \mathbf{W}_{t} \Vert d_{t}.
\end{equation}
According to \eqref{eq:decision_boudary}, data will be classified as class $t$ because $d_j<d_{t}$ when $\Vert \mathbf{W}_j \Vert = \Vert \mathbf{W}_{t} \Vert$. And our empirical observations show that head classes tend to have much larger margins  and  logits  than  tail  classes:
\begin{equation}
\label{eq:margin_relats}
\begin{split}
    & \bar{d}_1 >  \bar{d}_2 > \cdots > \bar{d}_K, \\
    & \bar{\eta}_1 >  \bar{\eta}_2 > \cdots > \bar{\eta}_K, \\
    & \text{if} \quad  n_1  >   n_2 > \cdots >  n_K,
\end{split}    
\end{equation}
where $\bar{d}_j$ and $\bar{\eta}_j$ are the average margin and logit of class $j$ after the standard training, respectively.
In detail, on sub-dataset $\mathcal{D}_j=\{(\mathbf{x}_i,y_i=j)\}_{i=1}^{n_j}$, $\bar{\eta}_j = \frac{1}{n_j} \eta_j$, $\bar{d}_j =\frac{\bar{\eta}_j}{\Vert \mathbf{W}_j \Vert} $. 

\subsection{\revision{Margin Calibration} (\revision{\method})}
To get the \revision{calibrated} logits, we propose \revision{\method} to \revision{calibrate} the margins after the standard training. Concretely speaking, we train a simple class-specific margin calibration model with the original margin fixed:
\begin{equation}
\label{eq:dis_rect}
\begin{split}
   \hat{d_j} = \omega_j \cdot d_j + \beta_j,
\end{split}    
\end{equation}
where \revision{$\omega_j$} and \revision{$\beta_j$} are learnable parameters for class $j$ and $j \in [1,K]$. \revision{It is worth noting that \revision{MARC} is extended from class $j$ to the whole dataset in practise}. In other words, \revision{MARC} only has $2K$ trainable parameters. Thus, the \revision{calibrated} logit is computed as:
\begin{equation}
\label{eq:dis_rec_score}
\begin{split}
   \Vert \mathbf{W}_j\Vert \hat{d_j} &=\Vert \mathbf{W}_j\Vert ( \omega_j \cdot d_j + \beta_j)\\
   & = \omega_j \cdot \Vert \mathbf{W}_j\Vert d_j + \beta_j \cdot \Vert \mathbf{W}_j\Vert \\
   & = \omega_j \cdot \eta_j  + \beta_j \cdot \Vert \mathbf{W}_j\Vert, \\
\end{split}    
\end{equation}
where $\eta_j$ is the initial fixed logit. Then, we can get the \revision{calibrated} prediction distribution:
\begin{equation}
    \label{eq:dis_rec_softmax}
    p(y=y_i|\mathbf{x}_i;\theta_r,\theta_c) = \frac{\exp(\omega_{y_i} \cdot \eta_{y_i}  + \beta_{y_i} \cdot \Vert \mathbf{W}_{y_i}\Vert)}{\sum_{j=1}^K  \exp(\omega_j \cdot \eta_j  + \beta_j \cdot \Vert \mathbf{W}_j\Vert)}.
\end{equation}
The training process of the margin calibration approach can be written with \emph{just three} lines of Pytorch codes as shown in Line 4-6 of Algorithm~\ref{alg:margin calibration function}.

\begin{algorithm}[ht]
\caption{The torch-like code for \revision{\method}.}
\label{alg:margin calibration function}
\begin{algorithmic}[1]
\STATE{Initialization of the margin calibration approach: \\
\footnotesize\texttt{omega=torch.nn.Parameter(torch.ones(1,K))}\\
\footnotesize\texttt{beta=torch.nn.Parameter(torch.zeros(1,K))}}
\STATE{\textbf{Input:} training data x, standard pre-trained neural network model.}
\STATE{\footnotesize\texttt{with torch.no\_grad():}}
\STATE{\quad \quad \footnotesize\texttt{w\_norm $=$ torch.norm(model.fc.weight, dim=1)}}
\STATE{\quad \quad \footnotesize\texttt{logit\_before $=$ model(x)}}
\STATE{\footnotesize\texttt{logit\_after $=$ omega * logit\_before + beta * w\_norm}}
\STATE{Compute loss and update parameters of omega and beta.}
\end{algorithmic}
\end{algorithm}

Furthermore, for more balanced gradients during training, we re-weight the loss as the previous work does~\citep{zhang2021distribution}. Finally, the loss for training the margin calibration approach is:
\begin{equation}
\label{eq:loss_dis_rec_softmax}
        \ell(\mathbf{x}_i, y_i;\tilde{\theta}_r,\tilde{\theta}_c,\omega,\beta) =
        - U_{y_{i}} \cdot \log \left(\frac{\exp \left(\omega_{y_i} \cdot \eta_{y_i}  + \beta_{y_i} \cdot \Vert \mathbf{W}_{y_i}\Vert \right)}{\sum_{j=1}^K  \exp(\omega_j \cdot \eta_j  + \beta_j \cdot \Vert \mathbf{W}_j\Vert)} \right),
\end{equation}
where $\tilde{\theta}_r $ and $\tilde{\theta}_c$ denote that these parameters are frozen during training. The weight for class $y_i$ is calculated as:
\begin{equation}
    \label{eq:weight}
         U_{y_i} = K \cdot \frac{(1/n_{y_i})^\gamma}{\sum_{j=1}^K(1/n_j)^\gamma },
\end{equation}
where $\gamma$ is a scale hyper-parameter. When $\gamma=0$, the weight for all classes is $1$, which means no re-weighting at all. 

To be more clear, the whole detailed training procedure including both standard training and margin calibration function training is demonstrated in Algorithm~\ref{alg:whole alg}. Lines 2-6 include the training procedure of the standard training using the instance-balanced sampling and the cross-entropy loss. Lines 7-11 contain the training process of our margin calibration function. It is \revision{worth} noting that in the second stage, parameters $\theta_r$ and $\theta_c$ are all fixed.

\subsection{Discussion}

We clarify the differences between \revision{MARC} and other learnable decision boundary adjustment methods in detail.
As shown in Table~\ref{tab:diff}, Decouple-cRT~\citep{kang2019decoupling} retrains the whole parameters of the classifier, while Decouple-LWS~\citep{kang2019decoupling} only adjusts the norm of weight vectors $\Vert \mathbf{W}_j \Vert$. Instead of adjusting the classifier head, DisAlign~\citep{zhang2021distribution} chooses to \revision{calibrate} the logit for each data point. But their calibration method is heuristic that simply adds the \revision{calibrated} logit and the original logit with a re-weighting scheme. To be more clear, the weighted sum of logits for DisAlign is $\sigma(\mathbf{z}_j) (\omega_j \eta_j + \beta) + (1-\sigma(\mathbf{z}_j))\eta_j$, where $\sigma(\cdot)$ is an instance-specific confidence function. However, different from previous methods, our \revision{\method} focuses on \revision{calibrat}ing the margin which we believe is the performance bottleneck of the long-tailed classifier.
\looseness=-1

\begin{table}[htbp]
\caption{The difference between \revision{MARC} and other decision boundary adjustment methods. $j\in [1,K]$ is class index.}
\label{tab:diff}
\centering
\resizebox{.6\textwidth}{!}{
\begin{tabular}{lc} \toprule
Method    &  calibration method \\\midrule
Decouple-cRT~\citep{kang2019decoupling} & retrain $\mathbf{W}_j, \mathbf{b}_j$  \\
Decouple-LWS~\citep{kang2019decoupling} & $  {\Vert \mathbf{W}_j \Vert} ^{1-\omega_j}$ \\
DisAlign~\citep{zhang2021distribution} & $\sigma(\mathbf{z}_j) (\omega_j \eta_j + \beta) + (1-\sigma(\mathbf{z}_j))\eta_j$\\
\revision{MARC} & $\omega_j \cdot d_j + \beta_j$  \\ \bottomrule
\end{tabular}}
\end{table}

\begin{algorithm}[t!]
\caption{The detailed training procedure including both standard training and margin calibration function training.}
\label{alg:whole alg}
\begin{algorithmic}[1]
\STATE{\textbf{Input:} The training dataset $\mathcal{D}=\{(\mathbf{x}_i,y_i)\}_{i=1}^n$, the parameters of the representation function $\theta_r$, the parameters of the classifier $\theta_c$, parameters of the margin calibration function $\omega$ and $\beta$, number of classes $K$ and the pre-defined scale hyper-parameter $\gamma$.}
\STATE{First stage: the standard training use the instance-balanced sampling and the cross entropy loss.}
\WHILE{not reach the maximum iteration}
\STATE{Use instance-balanced sampling to sample a batch of data $\mathcal{D}_s=\{(\mathbf{x}_i,y_i)\}_{i=1}^s$ from the training dataset $\mathcal{D}$, where $s$ is the batch size.}
\STATE{Compute the loss and update the model parameters.\\
$\ell(\mathcal{D}_s;\theta_r,\theta_c) = \frac{-1}{s} \sum_{i=1}^s \log \left(\frac{\exp (\eta_{y_i})}{\sum_{j=1}^K  \exp(\eta_j)}\right),$ where $\eta_j$ is classification score of class $j$. }
\ENDWHILE

\STATE{Second stage: \revision{Calibrate} the margins trained in the first stage.}
\WHILE{not reach the maximum iteration}
\STATE{Use instance-balanced sampling to sample a batch of data $\mathcal{D}_s=\{(\mathbf{x}_i,y_i)\}_{i=1}^s$ from the training dataset $\mathcal{D}$, where $s$ is the batch size.}
\STATE{Compute the loss and update the model parameters.\\
$\ell(\mathcal{D}_s;\tilde{\theta}_r,\tilde{\theta}_c,\omega,\beta) = \frac{1}{s} \sum_{i=1}^s( - U_{y_{i}} \cdot \log \left(\frac{\exp(\omega_{y_i} \cdot \eta_{y_i}  + \beta_{y_i} \cdot \Vert \mathbf{W}_{y_i}\Vert)}{\sum_{j=1}^K  \exp(\omega_j \cdot \eta_j  + \beta_j \cdot \Vert \mathbf{W}_j\Vert)} \right),$ where parameters with $\tilde{\cdot}$ are fixed during training and $U_j$ is calculated as shown in Eq.~\ref{eq:weight}. }
\ENDWHILE

\STATE{\textbf{Return:} Model parameters $\theta_r, \theta_c, \omega, \beta$.}

\end{algorithmic}
\end{algorithm}

\section{Experiments}

In this section, we conduct extensive experiments compared with the state-of-the-art methods to validate the effectiveness of \revision{MARC}.
First, we report performance on common benchmarks like CIFAR-LT~\citep{krizhevsky2009learning}, ImageNet-LT~\citep{liu2019large}, Places-LT~\citep{zhou2017places} and iNaturalist2018~\citep{van2018inaturalist}. The results of \revision{MARC} are competitive even though \revision{MARC} is simple. Then we conduct further analysis to explain the reason for the success of \revision{MARC}.

\subsection{Setup}

\paragraph{Datasets}
We follow the common evaluation protocol~\citep{liu2019large} and conduct experiments on CIFAR-10-LT~\citep{krizhevsky2009learning}, CIFAR-100-LT~\citep{krizhevsky2009learning}, ImageNet-LT~\citep{liu2019large}, Places-LT~\citep{zhou2017places} and iNaturalist2018~\citep{van2018inaturalist}. The imbalance factor used in CIFAR datasets is defined as $N_{max} / N_{min}$ where $N_{max}$ is the number of samples on the largest class and $N_{min}$ the smallest. We report CIFAR results with two different imbalance ratios: 100 and 200. For ImageNet-LT and Places-LT experiments, we further split classes into three sets: Many-shot (with more than 100 images), Medium-shot (with 20 to 100 images), and Few-shot (with less than 20 images).

\paragraph{Training Configuration}
For a fair comparison, our experiments are conducted under the most commonly used codebase of long-tailed studies: Open Long-Tailed Recognition (OLTR)~\citep{liu2019large}, using PyTorch~\citep{paszke2019pytorch} framework. The model structures used for CIFAR, ImageNet-LT, Places-LT and iNaturalist18 datasets are ResNet32, ResNeXt50, ResNet152 and ResNet50, respectively. The model for Places-LT is pre-trained on the full ImageNet-2012 dataset while models for other datasets are trained from scratch. For ImageNet-LT, Places-LT, and iNaturalist18, we train 90, 30, and 200 epochs in the first standard training stage; and 10, 10, and 30 epochs in the second margin calibration stage, with the batch size of 256, 128, and 256, respectively. For CIFAR-10-LT and CIFAR-100-LT, the models are trained for 13,000 iterations with a batch size of 512. We use the SGD optimizer with momentum 0.9 and weight decay $5e-4$ for all datasets except for iNaturalist18 where the weight decay is $1e-4$. In the standard training stage, we use a cosine learning rate schedule with an initial value of 0.05 for CIFAR and 0.1 for other datasets, which gradually decays to 0. In the margin calibration stage, we use a cosine learning rate schedule with an initial learning rate starting from 0.05 to 0 for all datasets. $\gamma$ is set to $1.2$ for all datasets. The hyper-parameters of compared methods follow their paper. \textit{For fairness, we use the same pre-trained model for decision boundary adjustment methods.}
\looseness=-1

\begin{minipage}{\textwidth}
\begin{minipage}[t!]{0.55\textwidth}
\centering
\makeatletter\def\@captype{table}\makeatother\caption{Accuracy on CIFAR-LT.}
\label{tab:cifar_res}
\resizebox{\textwidth}{!}{
\begin{tabular}{lcccc}
\toprule
\multicolumn{1}{l}{Dataset} & \multicolumn{2}{c}{CIFAR-10-LT} & \multicolumn{2}{c}{CIFAR-100-LT} \\ \midrule
Imbalance Factor       & \;\;\;100\;\;\;     & \;\;\;200\;\;\;    & \;\;\;100\;\;\;     & \;\;\;200\;\;\; \\ \midrule
Softmax                & 78.7          & 74.4         & 45.3          & 41.0          \\\midrule
 Data Re-sampling                        &   &      &         &           \\
Class Balanced Sampling (CBS)                &77.8           &68.3           &42.6           &37.8       \\ \midrule
Loss Function Engineering        &   &      &         &           \\
Class Balanced Weighting (CBW)                &78.6            &72.5           &42.3           &36.7       \\
Class Balanced Loss~\citep{cui2019class}     &78.2           &72.6           &44.6           &39.9       \\
Focal Loss~\citep{focal}             & 77.1          & 71.8         & 43.8          & 40.2          \\
LADE~\citep{hong2021disentangling}                   &      81.8         & 76.9          & 45.4          &     43.6          \\ 
LDAM~\citep{LDAM}                   & 78.9          & 73.6         & 46.1          & 41.3          \\
Equalization Loss~\citep{equal}       &78.5           &74.6           &47.4           &43.3       \\
Balanced Softmax~\citep{ren2020balanced}       & 83.1          & 79.0         & 50.3          & 45.9         \\ \midrule
Decision Boundary Adjustment        &   &      &         &           \\
DisAlign~\citep{zhang2021distribution}              & 78.0         & 71.2          & 49.1          &  43.6    \\
Decouple-cRT~\citep{kang2019decoupling}           & 82.0          & 76.6         & 50.0          & 44.5          \\
Decouple-LWS~\citep{kang2019decoupling}           & 83.7          & 78.1         & 50.5          & 45.3          \\
 \midrule
Others        &   &      &         &           \\
BBN~\citep{zhou2020bbn}                    & 79.8          &      -     & 42.6          &  -   \\ 
Hybrid-SC~\citep{wang2021contrastive}        & 81.4          &   -     & 46.7          & - \\ \midrule
MARC                   & \textbf{85.3}          & \textbf{81.1}         & \textbf{50.8}         & \textbf{47.4} \\\bottomrule
\end{tabular}
}
\end{minipage}
\begin{minipage}[t!]{0.45\textwidth}
\centering
\makeatletter\def\@captype{table}\makeatother\caption{Accuracy on iNaturalist-LT.}
\label{tab:inaturalist_res}
\resizebox{\textwidth}{!}{
\begin{tabular}{lc} \toprule
Method              & Top-1 Accuracy \\ \midrule
Softmax & 65.0 \\ \midrule
Loss Function Engineering & \\
Class Balanced Loss~\citep{cui2019class}   & 61.1\\
LDAM~\citep{cao2019learning}       & 64.6\\
Balanced Softmax~\citep{ren2020balanced} & 69.8\\
LADE~\citep{hong2021disentangling} & 70.0\\ \midrule
Decision Boundary Adjustment        &         \\
Decouple-$\pi$-norm~\citep{kang2019decoupling} & 69.3\\
Decouple-LWS~\citep{kang2019decoupling}   & 69.5\\
DisAlign~\citep{zhang2021distribution} & \underline{70.3}\\ \midrule
Others & \\
Casual Norm~\citep{tang2020long} & 63.9 \\ 
Hybrid-SC~\citep{wang2021contrastive}      & 68.1  \\\midrule
MARC    & \textbf{70.4}\\\bottomrule
\end{tabular}
}
\end{minipage}
\end{minipage}

\subsection{Comparison with previous methods}
In this section, we compare the performance of \revision{MARC} to other recent works. We select some recent methods from each of the following four categories for comparison: data re-sampling, loss function engineering, decision boundary adjustment, and others. The standard training with the cross-entropy loss and instance balance sampling is called Softmax in our results.

\paragraph{CIFAR}
Table~\ref{tab:cifar_res} presents results for CIFAR-10-LT and CIFAR-100-LT. \revision{MARC} outperforms all other methods in CIFAR-LT. Compared with other decision boundary adjustment methods, \revision{MARC} shows favorable results. The accuracy of \revision{MARC} outruns Decouple-LWS $1.6\%$, $3\%$, $0.3\%$ and $1.9\%$ on CIFAR-10-LT(100), CIFAR-10-LT(200), CIFAR-100-LT(100) and CIFAR-100-LT(200) respectively, where ($\cdot$) denotes the imbalance factor. In addition, \revision{MARC} outperforms all data re-sampling and loss function engineering methods that my need laborious hyper-parameter. The performance of well-designed networks such as BBN and Hybrid-SC are also not as good as that of \revision{MARC}.

\paragraph{ImageNet-LT}
We further evaluate \revision{MARC} on the ImageNet-LT dataset. As Table~\ref{tab:imagenet_res} shows, \revision{MARC} is better than all loss function engineering methods. Compared with LADE, although our overall accuracy is $0.4\%$ higher, the accuracy on the few-shot classes is $5.4\%$ higher. The few-shot accuracy and overall accuracy of \revision{MARC} are $1.9\%$ and $0.1\%$ higher than DisAlign respectively. Our results are quite surprising
considering the simplicity of \revision{MARC} (See next section for time comparison).

\begin{minipage}{\textwidth}
\begin{minipage}[t!]{0.45\textwidth}
\centering
\makeatletter\def\@captype{table}\makeatother\caption{Accuracy on ImageNet-LT.}
\label{tab:imagenet_res}
\resizebox{\textwidth}{!}{
\setlength{\tabcolsep}{0.5mm}
\begin{tabular}{lcccc} \toprule
Method              & Many & Medium & Few  & Overall \\ \midrule
Softmax             & \underline{65.1} & 35.7   & 6.6  & 43.1    \\ \midrule
Loss Function Engineering        &   &      &         &           \\
Focal Loss~\citep{focal}          & 64.3 & 37.1   & 8.2  & 43.7    \\
Seesaw~\citep{wang2021seesaw}              & \textbf{67.1} & 45.2   & 21.4 & 50.4    \\
Balanced Softmax~\citep{ren2020balanced}    & 62.2 & 48.8   & 29.8 & 51.4     \\
LADE~\citep{hong2021disentangling}                & 62.3 & 49.3   & 31.2 & 51.9    \\
\midrule
Decision Boundary Adjustment        &   &      &         &           \\
Decouple-$\pi$-norm~\citep{kang2019decoupling} & 59.1 & 46.9   & 30.7 & 49.4    \\
Decouple-cRT~\citep{kang2019decoupling}        & 61.8 & 46.2   & 27.4 & 49.6    \\
Decouple-LWS~\citep{kang2019decoupling}        & 60.2 & 47.2   & 30.3 & 49.9    \\
DisAlign~\citep{zhang2021distribution}            & 60.8 & \textbf{50.4}   & \underline{34.7} & \underline{52.2}    \\ \midrule
Others        &   &      &         &           \\
OLTR~\citep{liu2019large}                 & 51.0 & 40.8   & 20.8 & 41.9    \\
Causal Norm~\citep{tang2020long}           & 62.7 & 48.8   & 31.6 & 51.8    \\ \midrule
MARC                & 60.4 & \underline{50.3}   & \textbf{36.6} & \textbf{52.3}     \\ \bottomrule
\end{tabular}
}
\end{minipage}
\begin{minipage}[t!]{0.45\textwidth}
\centering
\makeatletter\def\@captype{table}\makeatother\caption{Accuracy on Places-LT}
\label{tab:places_res}
\resizebox{\textwidth}{!}{
\begin{tabular}{lcccc} \toprule
Method              & Many & Medium & Few  & Overall \\ \midrule
Softmax             & 46.4 & 27.9   & 12.5 & 31.5    \\\midrule
Loss Function Engineering        &   &      &         &           \\
Focal Loss~\citep{focal}          & 41.1 & 34.8   & 22.4 & 34.6    \\
Balanced Softmax~\citep{ren2020balanced}    & 42.0  &38.0     &17.2 & 35.4  \\
LADE~\citep{hong2021disentangling}               & \underline{42.8} & 39.0   & 31.2 & \textbf{38.8}    \\ \midrule
Decision Boundary Adjustment        &   &      &         &           \\
Decouple-LWS~\citep{kang2019decoupling}        & 40.6 & 39.1   & 28.6 & 37.6    \\
Decouple-$\pi$-norm~\citep{kang2019decoupling} & 37.8 & \textbf{40.7}   & 31.8 & 37.9    \\
DisAlign~\citep{zhang2021distribution}             & 40.0 &	39.6 &32.3 	& 38.3\\ \midrule
Others        &   &      &         &           \\ 
OLTR~\citep{liu2019large}           & \textbf{44.7} & 37.0   & 25.3 & 35.9    \\
Causal Norm~\citep{tang2020long}      & 23.8 & 35.8   &\textbf{40.4} & 32.4    \\
 \midrule
MARC                & 39.9 & \underline{39.8} 	& \underline{32.6} & \underline{38.4} \\\bottomrule
\end{tabular}
}
\end{minipage}
\end{minipage}

\paragraph{Places-LT}
For the Places-LT dataset, \revision{MARC} achieves better performance than other decision boundary adjustment methods. Though our overall accuracy is lower than LADE,  our few-shot accuracy is still $1.4\%$ higher than LADE. Though \revision{MARC} is not the best in Places-LT, the results of \revision{MARC} are still competitive compared with other methods.

\paragraph{iNaturalist-LT}
Finally, we present the Top-1 accuracy results for iNaturalist-LT dataset in Table~\ref{tab:inaturalist_res}. We can observe a similar trend that our proposed method wins all the existing approaches and surpasses DisAlign by 0.1\% absolute improvement. 


\subsection{Effectiveness validation}

\paragraph{Comparison of the trainable parameters of decision boundary adjustment methods}
As shown in Table~\ref{tab:parameters}, \revision{MARC} achieves the best performance among other compared methods on CIFAR-100-LT(200) and ImageNet-LT. Even though our trainable parameters are more than Decouple-LWS, our performance is better. Besides, it is surprising that \revision{MARC} obtains such a favorable performance with only so few parameters.

\begin{table}[htbp]
\caption{\textbf{Comparison of the trainable parameters (\#Param.) of the learnable decision boundary adjustment methods.} $p$ is the feature dimension and $K$ is the number of classes (ResNeXt50 for ImageNet-LT: $p=2048, K=1000$).}
\label{tab:parameters}
\centering
\resizebox{0.45 \textwidth}{!}{
\begin{tabular}{lccc} \toprule
Method    &   CIFAR-100-LT & ImageNet-LT &\#Param. \\\midrule
Decouple-cRT & 44.5 & 49.6 &  $pK + K$ \\
Decouple-LWS & \underline{45.3} & 49.9& $K$ \\
DisAlign & 43.6 & \underline{52.2} & $p + 2K$ \\
\revision{MARC} & \textbf{47.4} & \textbf{52.3} & $2 K$ \\ \bottomrule
\end{tabular}}
\end{table}

\paragraph{Effects of different standard pre-trained model}
We use different standard pre-trained models on CIFAR-100-LT(200) to explore their effects. As Table~\ref{tab:pretrain_effect} illustrates, as the pre-trained dataset gets more balanced, the performance of our margin calibration method gets better. This shows that when comparing decision boundary methods, using the same standard training model is very important for fairness. And the codebase will also affect the standard training. \textit{This is the reason why the result of DisAlign in our paper is inconsistent with the original paper since we cannot get the same standard pre-trained model they used}. So instead, we use our own standard pre-trained model for \revision{MARC} and DisAlign for fairness.
Table~\ref{tab:pretrain_effect}
also demonstrates that the margin calibration method can achieve better performance when given better representations. So our future works include how to get better representations.
\looseness=-1

\begin{table}[htbp]
\caption{Top-1 accuracy on CIFAR-100-LT(200) with different standard pre-trained models.}
\label{tab:pretrain_effect}
\centering
\resizebox{0.45 \textwidth}{!}{
\begin{tabular}{lc} \toprule
Standard pre-trained  dataset           & Top-1 Accuracy \\ \midrule
 CIFAR-100-LT(200) & 47.1 \\ 
CIFAR-100-LT(100) & 50.7 \\ 
CIFAR-100-LT(50) & 54.5 \\ \bottomrule
\end{tabular}}
\end{table}

\begin{figure}[b!]
     \centering
     \subfigure[Logits]{
         \includegraphics[width=0.20\linewidth]{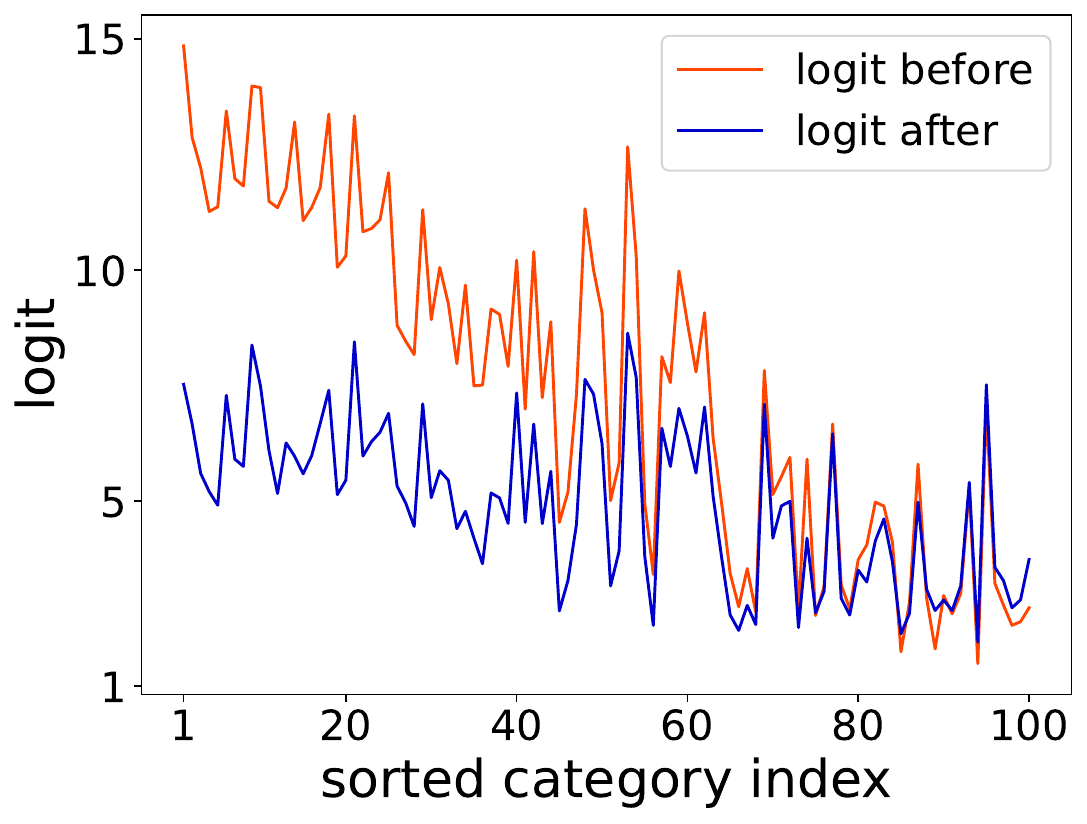}
         \label{fig:logit cifar100}}
     \subfigure[Margins]{
         \includegraphics[width=0.20\linewidth]{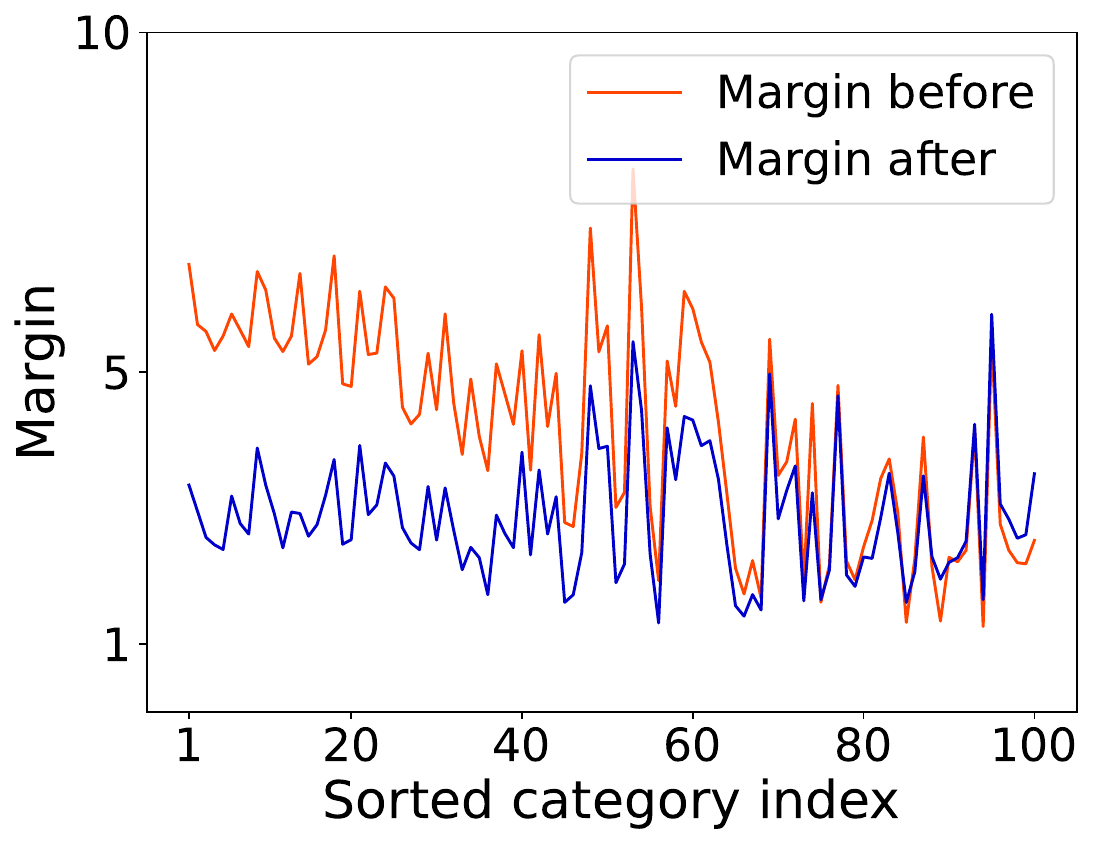}
         \label{fig:margin cifar100}}
     \rulesep
     \subfigure[Standard training.]{
         \includegraphics[width=0.20\linewidth]{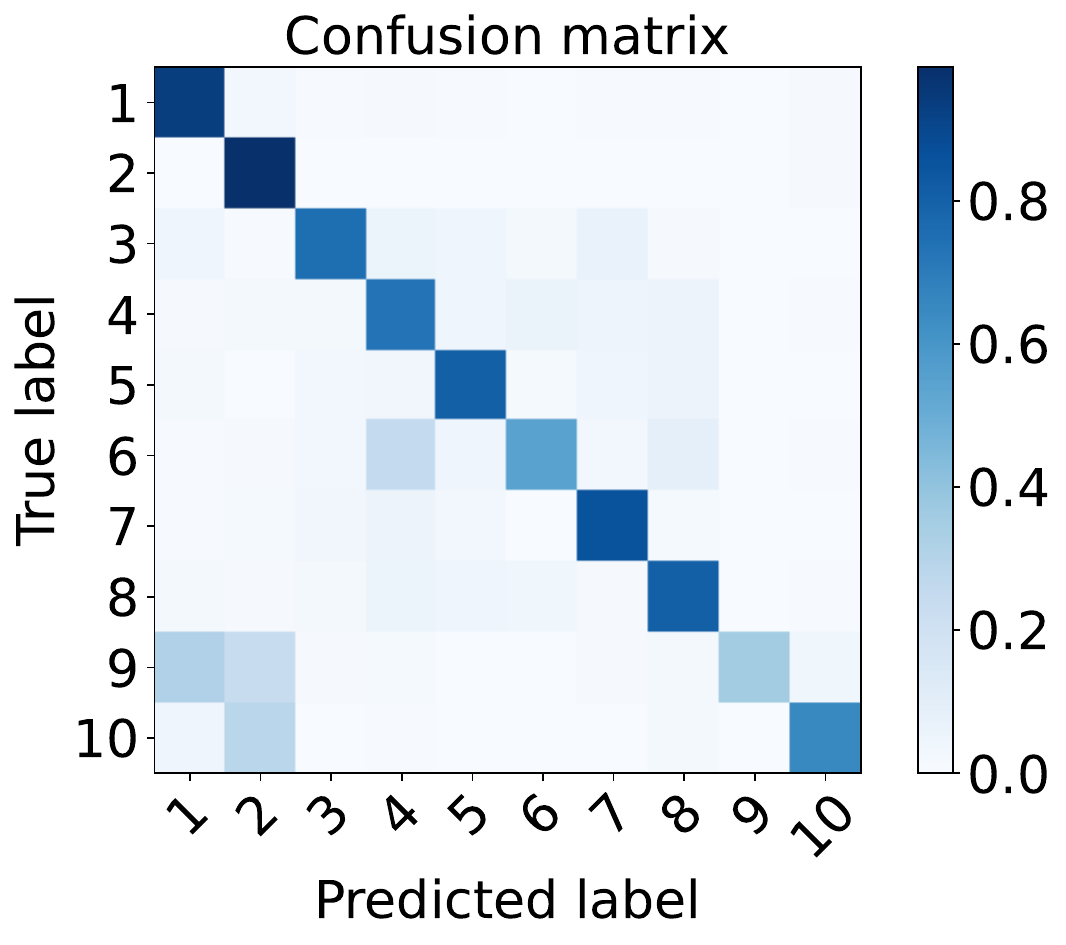}
         \label{fig:confusion matrix cifar10 softmax}}
     \subfigure[MARC]{
         \includegraphics[width=0.20\linewidth]{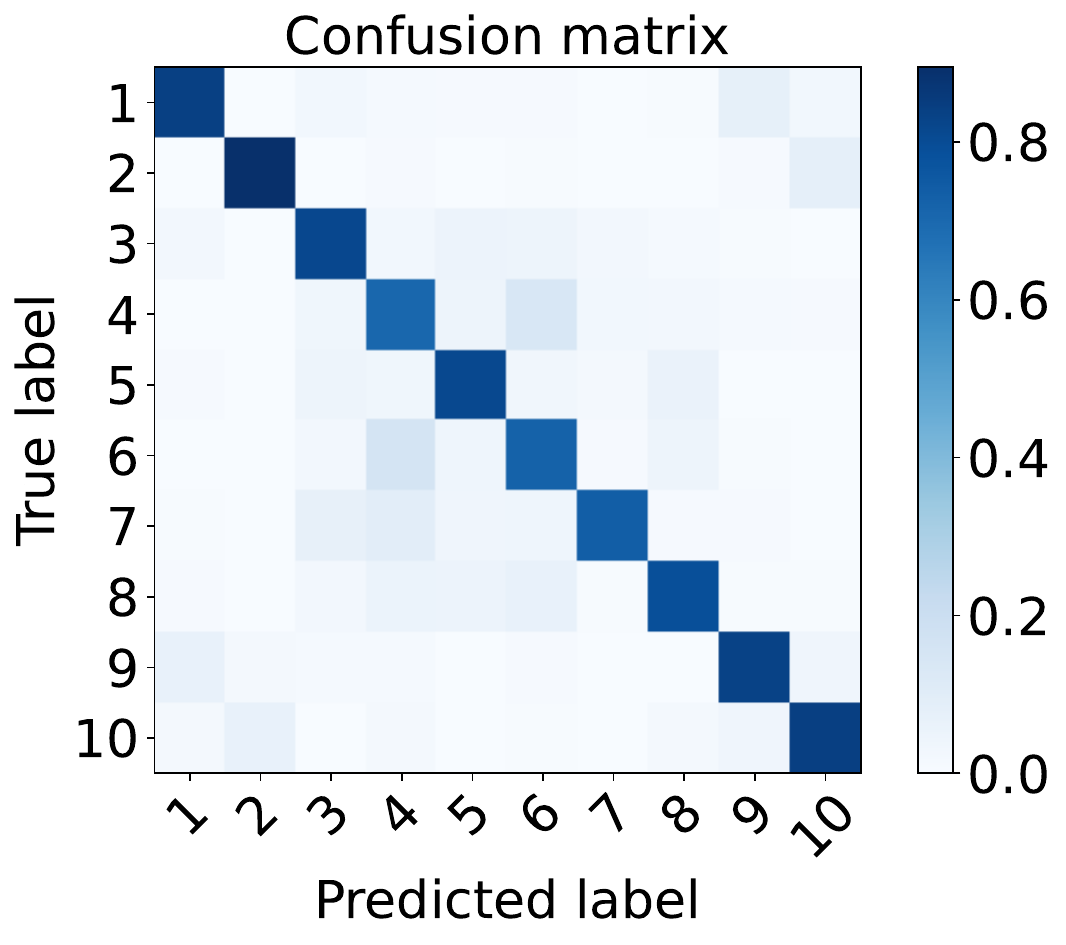}
         \label{fig:confusion matrix cifar10 our}}
        \caption{(a) and (b): \textbf{Logits and margins of CIFAR-100-LT(200)}. Here, \revision{the logit and margin represent the average logit and margin of each class}, \textit{before} refers to the standard training results and \textit{after} refers to the results after our calibration method. The class \revision{indices} are sorted by the number of samples (Head to tail). (c) and (d): \textbf{Confusion matrix of the standard training and \revision{\method} on long-tailed CIFAR-10-LT(200).} The fading color of diagonal elements refers to the disparity of the accuracy.}
        \label{fig:logits and margins cifar100}
\end{figure}

\subsection{Further analysis}

In this section, we conduct different experiments for further analysis. To be concrete, we empirically show that \revision{MARC} can achieve more balanced margins and logits compared with DisAlign. Moreover, the class-wise accuracy of \revision{MARC} is much better than the standard training baseline model on CIFAR-LT, which indicates that we can alleviate the imbalanced prediction problem and reduce the performance gap between the head and the tail classes.

\begin{figure*}[t!]
     \centering
         \subfigure[Margins of DisAlign.]{
         \includegraphics[width=0.20\linewidth]{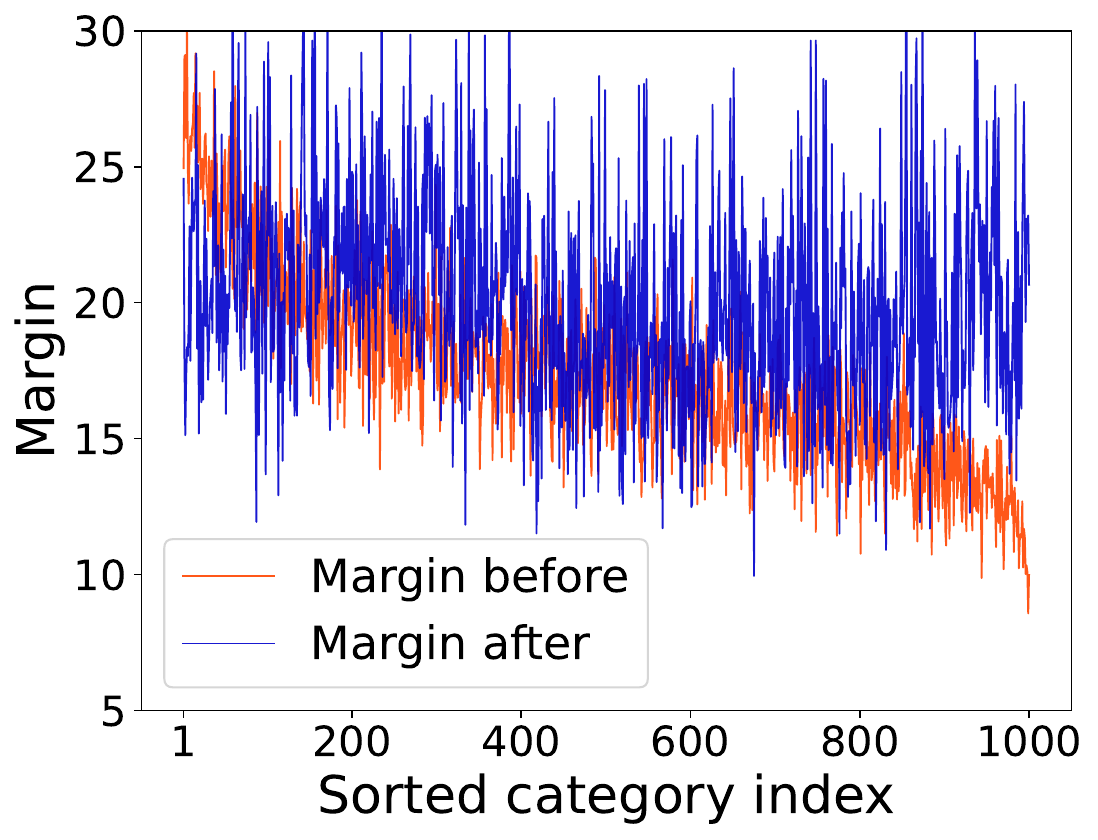}
         \label{fig:margin imagenet disalign}}
     \subfigure[Logits of DisAlign.]{
         \includegraphics[width=0.20\linewidth]{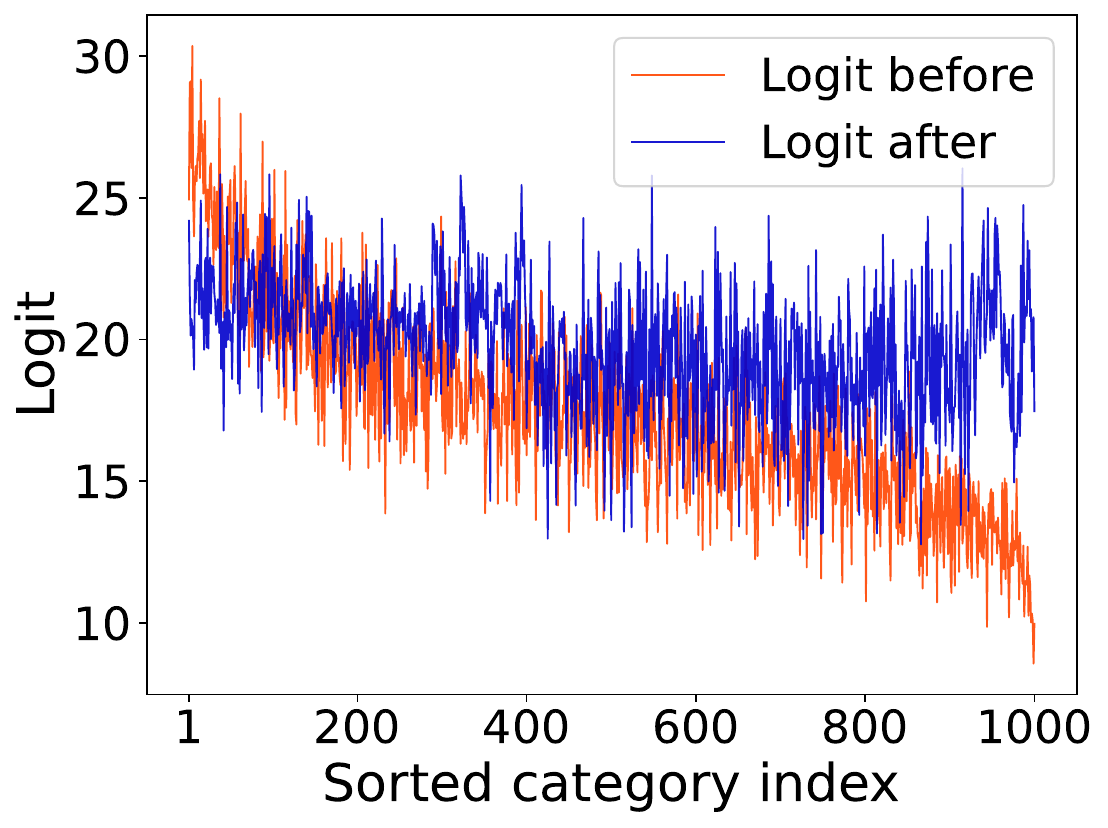}
         \label{fig:logit imagenet disalign}}
     \subfigure[Margins of \revision{MARC}.]{
         \includegraphics[width=0.20\linewidth]{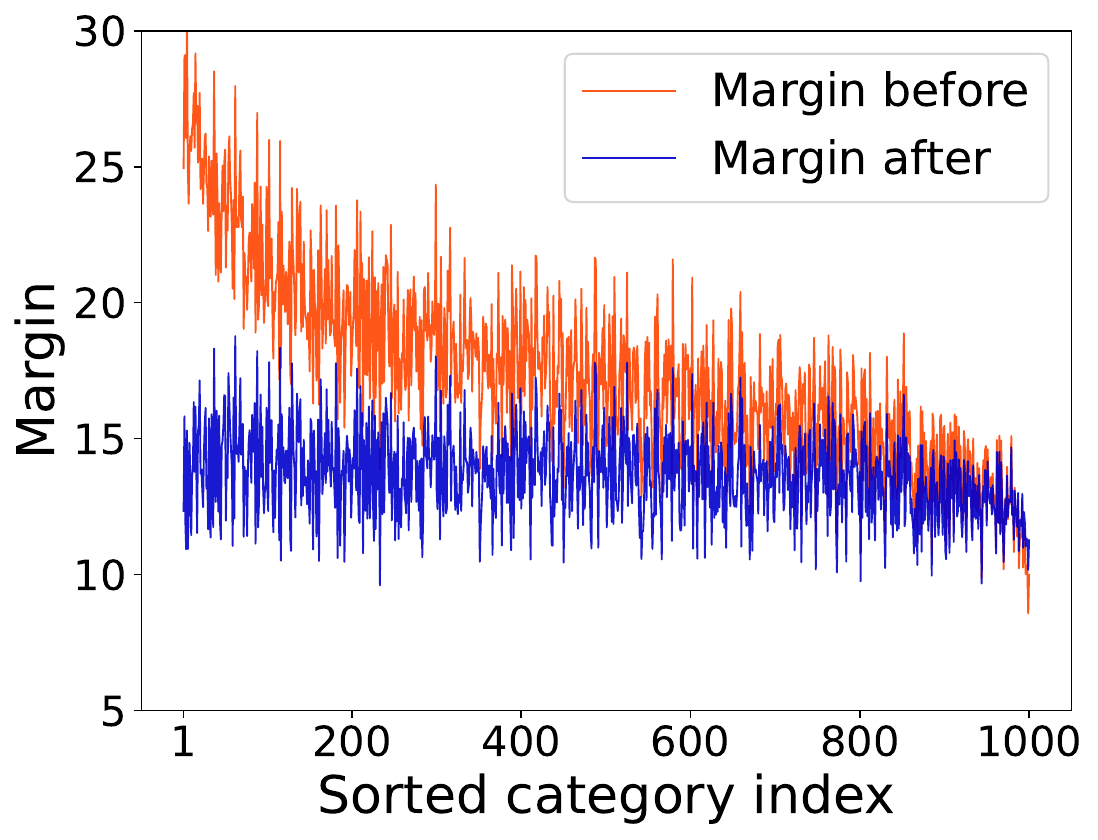}
         \label{fig:margin imagenet}}
     \subfigure[Logits of \revision{MARC}.]{
         \includegraphics[width=0.20\linewidth]{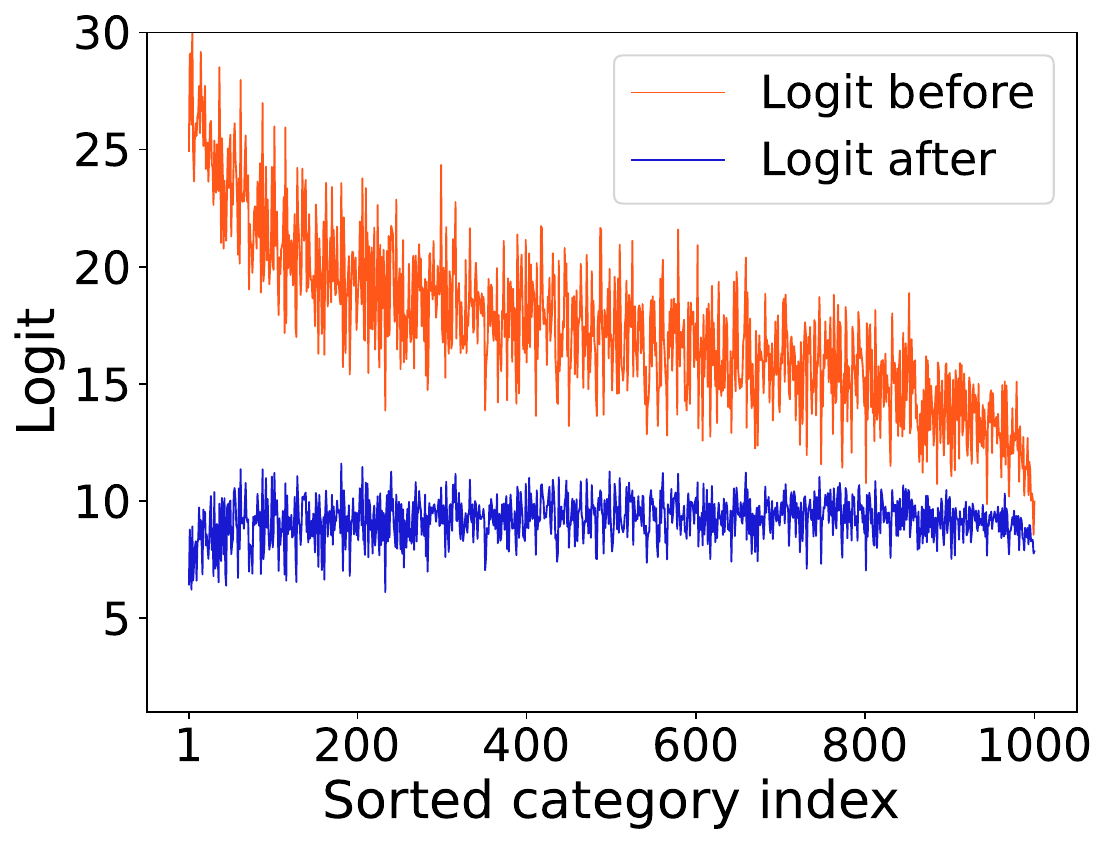}
         \label{fig:logit imagenet}}
        \caption{\textbf{The values of margins and logits for each class on the ImageNet-LT dataset.} \revision{The logit and margin represent the average logit and margin of each class}. The class \revision{indices} are sorted by the number of samples (Head to tail). \textit{Before} refers to the standard training results and \textit{after} refers to the results after using the decision boundary adjustment method.}
        \label{fig:logits and margins imagenet}
\end{figure*}

\paragraph{Visualization of the margin and logit}
In this subsection, we visualize the values of margins and logits for each class to show the effect of \revision{MARC}. As illustrated both in Figure~\ref{fig:logits and margins cifar100}, Figure~\ref{fig:margin imagenet} and Figure~\ref{fig:logit imagenet}, before margin calibration, the margins and the logits are \revision{uncalibrated}, i.e. the head classes tend to have much larger margins and logits than tail classes. We believe the bias in margins and logits will lead to imperfect predictions in long-tailed visual recognition. The margins and logits become more balanced after the margin calibration. This result proves that we can get better predictions by calibrating the margin. Moreover, as shown in Figure~\ref{fig:margin imagenet disalign} and Figure~\ref{fig:logit imagenet disalign}, \revision{MARC} will obtain more balanced margins and gradients than DisAlign. The instability of DisAlign may be caused by their heuristic design of the combination of the calibrated logits and the origin logits.

\begin{figure}[t!]
     \centering
     \subfigure[CIFAR-100-LT(200).]{
         \includegraphics[width=0.28\linewidth]{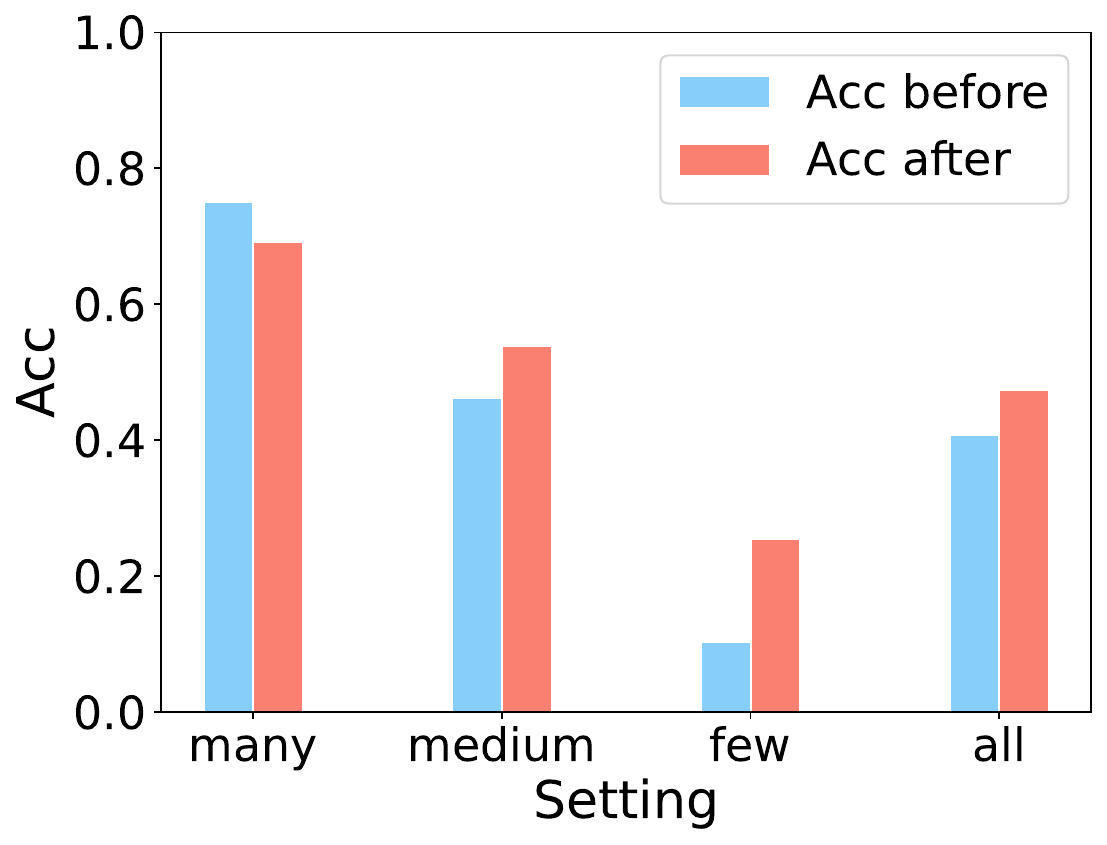}
         \label{fig:cifar100 acc}}
      \subfigure[F1 scores on CIFAR-LT.]{
         \includegraphics[width=0.28\linewidth]{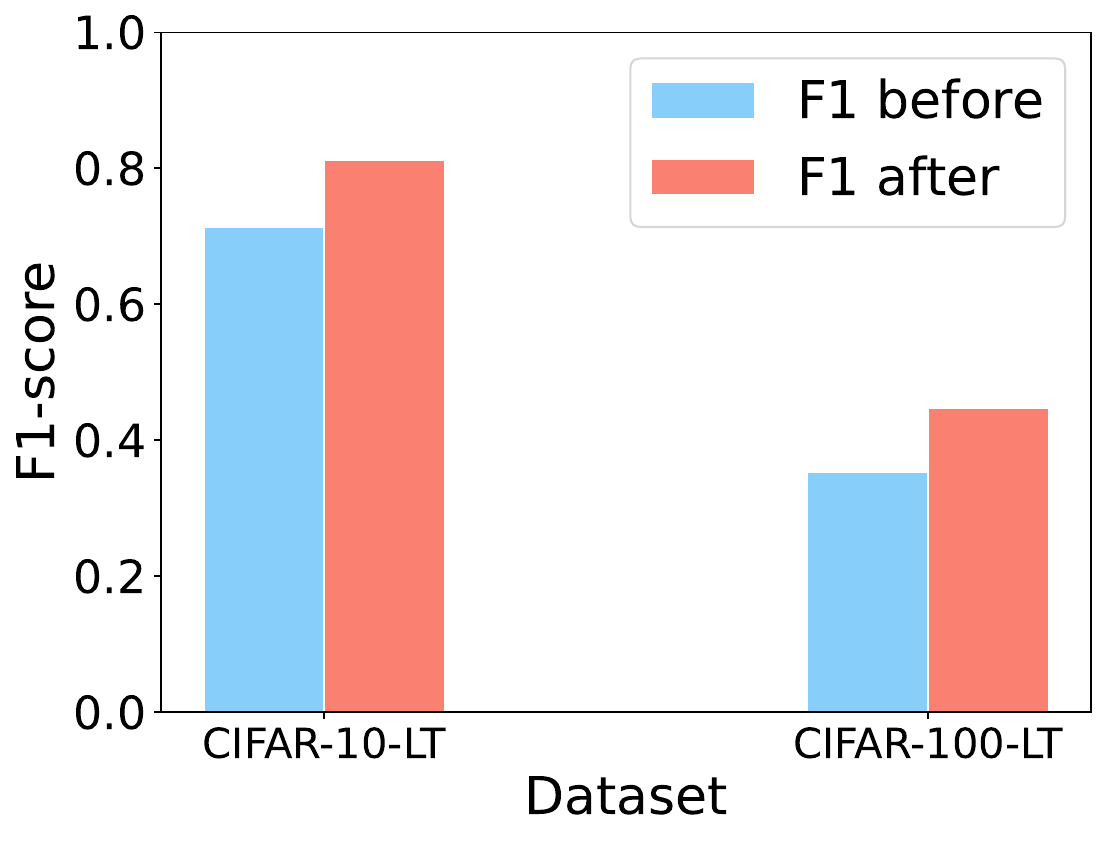}
         \label{fig:cifar10 f1}}
        \caption{The detailed performance of CIFAR-100-LT(200). \textit{Before} refers to the standard training results and \textit{after} refers to the results after \revision{MARC}.}
        \label{fig:cifar_acc}
\end{figure}

\paragraph{Class-wise performance on CIFAR-LT}

As we can see in \figurename~\ref{fig:confusion matrix cifar10 softmax} and \ref{fig:confusion matrix cifar10 our}, after our margin \revision{calibration} method, the performance on tail classes is improved while that on head classes is not severely affected. More intuitively, \figurename~\ref{fig:cifar_acc} shows the class-wise performance. The accuracy of the tail class is much higher than that of the head class. The performance degradation on head classes may be caused by the false positive predictions on head classes, i.e., the standard training method tends to classify tail classes as head classes, resulting in high accuracy on the head classes. The bad performance on tail classes when using standard training also proves this. In addition, the overall accuracy and F1-score show that \revision{\method} alleviates the \revision{uncalibrated} prediction problem.
\begin{wrapfigure}{r}{0.4\textwidth}
\centering
\includegraphics[width=0.3\textwidth]{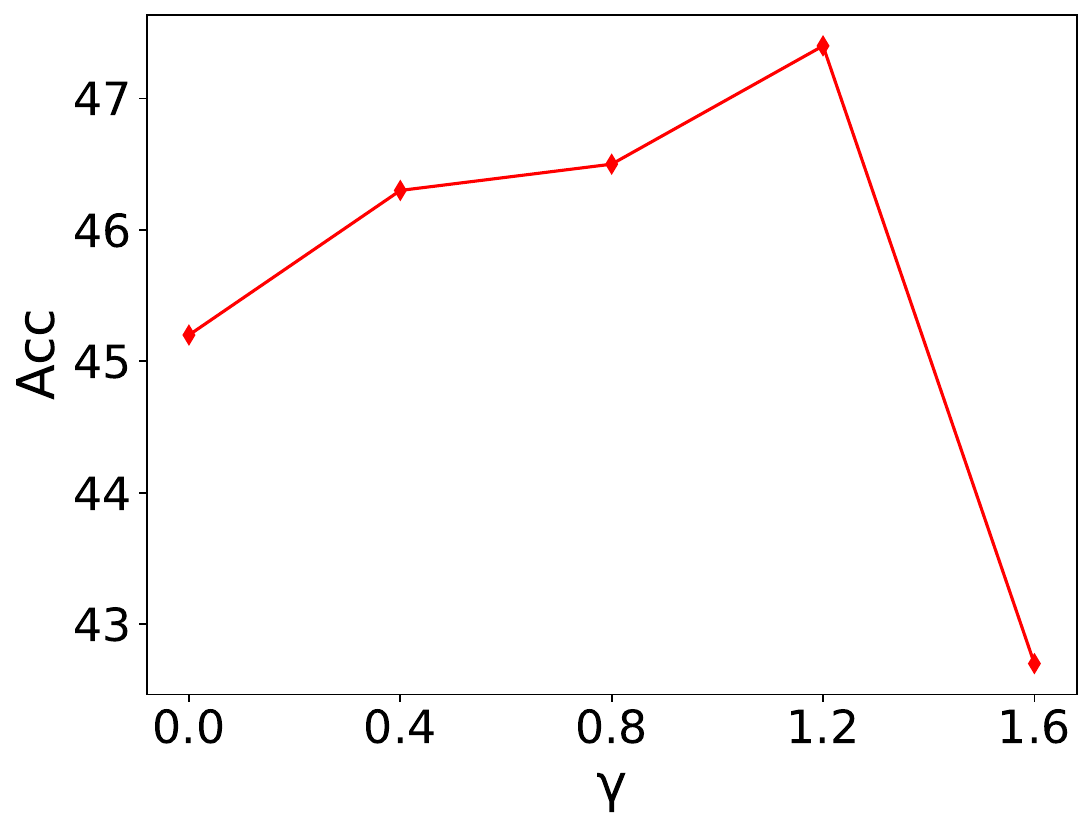}
\caption{Accuracy on CIFAR-100-LT(200) with different $\gamma$.}
\vspace{-.2in}
\label{fig:gamma}
\end{wrapfigure}



\paragraph{Effects of different $\gamma$}
To explore the effect of different $\gamma$, we also conduct experiments and visualize the performances on all CIFAR-100-LT(200). The results are shown in Figure~\ref{fig:gamma}.
We can observe that compared $1.2$ is the best compared with other values. $\gamma$ can not be too large since in this way the weight for head classes is too small. It is \revision{worth} noting that \revision{MARC} also achieves $45.2\%$ accuracy when $\gamma$ is $0$. This means \revision{MARC} still works even if we do not use any loss re-weighting techniques in the second stage. For other datasets, we directly use $1.2$ for $\gamma$.

\section{Conclusions}
This paper studied the long-tailed visual recognition problem.
Specifically, we found that head classes tend to have much larger margins and logits than tail classes. Motivated by our findings, we proposed a margin calibration function with only $2K$ learnable parameters to obtain the balanced logits in long-tailed visual recognition. Even though our method is very simple to implement, \revision{extensive experiments show that \revision{MARC} achieves favorable results compared with previous methods without altering the model representation. We hope that our study on logits and margins could provide experience for joint optimization of the model representation and margin calibration}.In the future, we aim to develop a unified theory to better support our algorithm design and apply this algorithm to more long-tailed applications.

\acks{We would like to thank anonymous reviewers for their insightful suggestions to help improve the paper. This study was supported by JSPS KAKENHI Grand Number JP22K12069.}

\bibliography{acml22}

\end{document}